\documentclass[lettersize,journal]{IEEEtran}
\usepackage{amsmath,amsfonts}
\usepackage{algorithmic}
\usepackage{algorithm}
\usepackage{array}
\usepackage[caption=false,font=normalsize,labelfont=sf,textfont=sf]{subfig}
\usepackage{textcomp}
\usepackage{booktabs}
\usepackage{stfloats}
\usepackage{url}
\usepackage{verbatim}
\usepackage{graphicx}
\usepackage{cite}
\usepackage{xcolor}
\usepackage{todonotes}
\usepackage[table]{xcolor}
\usepackage{siunitx}  
\begin{document}

% \title{Model Selection Methodology (MSM)}
\title{Mirror, Mirror on the Wall -- Which is the Best Model of Them All?}

\author{Dina Sayed \qquad Heiko Schuldt \\
Databases and Information Systems Research Group \\
University of Basel, Switzerland \\
\{dina.awad\textbar heiko.schuldt\}@unibas.ch}
\maketitle
\begingroup
\renewcommand\thefootnote{}
\footnotetext{
This work has been submitted to the IEEE for possible publication. Copyright may be transferred without notice, after which this version may no longer be accessible.}
\addtocounter{footnote}{-1}
\endgroup

% The paper headers
\markboth{
%Journal of \LaTeX\ Class Files,~Vol.~14, No.~8, August~2021}%
Sayed \& Schuldt: Mirror, Mirror on the Wall -- Which is the Best Model of Them All?}{}
% An Exploratory Study of Benchmarks and Leaderboards for Medical LLMs}{}

\maketitle

\begin{abstract}
Large Language Models (LLMs) have become one of the most transformative tools across many applications, as they have significantly boosted productivity and achieved impressive results in various domains such as finance, healthcare, education, telecommunications, and law, among others. Typically, state-of-the-art (SOTA) foundation models are developed by large corporations based on large data collections and substantial computational and financial resources required to pretrain such models from scratch. These foundation models then serve as the basis for further development and domain adaptation for specific use cases or tasks. However, given the dynamic and fast-paced nature of launching new foundation models, the %decision-making 
process of selecting the most suitable model for a particular use case, application, or domain becomes increasingly complex.
We argue that there are two main dimensions that need to be taken into consideration when selecting a model for further training: a qualitative dimension (which model is best suited for a task -- based on information, for instance, taken from model cards~\cite{mitchell2019model}) and a quantitative dimension (which is the best performing model).  The quantitative performance of models is assessed through leaderboards, which rank models based on standardized benchmarks and provide a consistent framework for comparing different LLMs. In this work, we address the analysis of the quantitative dimension by exploring the current leaderboards and benchmarks. To illustrate this analysis, we focus on the medical domain as a case study, demonstrating the evolution, current landscape, and practical significance of this quantitative evaluation dimension. 
%For the qualitative dimension, we examine additional factors that contribute to establishing trust in models. 
Finally, we propose a \textit{Model Selection Methodology (MSM)}, a systematic approach designed to guide the navigation, prioritization, and selection of the model that best aligns with a given use case.

\end{abstract}


\section{Introduction}
\label{sec:intro}
\IEEEPARstart{C}{onsider} % \textcolor{blue}{ 
a scenario in which a professional seeks to perform a task with greater efficiency; for example, a medical expert may wish to produce a concise summary of a case or generate a structured report. In previous decades, such tasks often required significant time and effort. However, with the advancement of natural language processing, and in particular the emergence of large language models (LLMs), these tasks have become more manageable and can now be completed in a fast and effective manner. LLMs have transformed numerous domains and are considered indispensable tools across a wide range of professions. In response to this growth, many organizations and research institutions are releasing their own LLMs at a rapid pace, in ever increasing quality. 

This raises the key question: \textit{Which model is most appropriate and best performing for a given task and set of requirements?}
Selecting a suitable model for subsequent fine-tuning necessitates the evaluation of two primary criteria: a \emph{quantitative} dimension and a \emph{qualitative} dimension, as follows:

\begin{itemize}
\item Qualitative dimension: involves \textit{human-centric evaluation %\todo{Do we need 'beyond numerical scores'´}
of a model's functionality}.  % beyond numerical scores. 
This includes carefully reviewing the model card for the authors’ reported limitations and intended use cases~\cite{mitchell2019model}. It can also include methods such as Report Card, which focus on generating natural language reports that analyze the model’s specificity, faithfulness, and interpretability~\cite{yang2024report,murahari2023qualeval}.

\item Quantitative dimension: focuses on answering \textit{how well a model performs numerically}. We examine the performance aspect by reviewing current model rankings across various leaderboards and benchmarks. These rankings provide a standardized quantitative measure of how different LLMs perform, enabling consistent comparisons across models. 

\end{itemize}

In this paper, we mainly address the analysis of the quantitative dimension. 
To further illustrate this analysis, we focus on the medical domain as an example, which helps to highlight the benefits of using such metrics, shed light on the challenges of maintaining these leaderboards, and emphasize that this cannot be the sole method for model selection.
Finally, based on our analysis, we propose a systematic approach for model selection.
% introduce of these two dimensions, we introduce a Model Selection Method (MSM), a systematic approach designed to assist in the navigation, prioritization, and selection of the model that best fits a given use case.

The contributions of this paper are threefold: First, we explore the quantitative dimension by introducing various leaderboards and delving deeper into the medical domain as an example of a domain-specific case, navigating through medical leaderboards and benchmarks. Second, we shed light on the insights that can be derived from community endorsements of different models. Third, we propose the Model Selection Methodology (MSM), an intuitive and systematic approach that supports the selection and shortlisting of the most suitable models for specific use cases.
%we examine the qualitative dimension by shedding light on community endorsement of different models. Finally, we propose the Model Selection Methodology (MSM), an intuitive and systematic approach that supports the selection and shortlisting of the most suitable models for specific use cases.

The remainder of this paper is structured as follows: 
Section~\ref{sec:quantD} introduces various leaderboards as the quantitative dimension and Section~\ref{sec:medleaderboards} surveys existing medical leaderboards. Section~\ref{sec:benchmarks} illustrates commonly used benchmarks in the medical domain and the discussion in Section~\ref{sec:lbdiscussion} outlines how to address the most pertinent challenges in LLM leaderboards. Section~\ref{sec:commu} explores the community voting metrics for models. Section~\ref{sec:msm} proposes the Model Selection Methodology (MSM) to help in selecting the most suitable models using a structured approach. Section~\ref{sec:discussion} discusses the need for standardizing the model selection process. Finally, Section~\ref{sec:conc} concludes the paper.

\section{Leaderboards and Benchmarks -- An Overview}
\label{sec:quantD}
LLM leaderboards have been introduced as a mechanism to provide systematic comparative insights into various LLMs.
LLM leaderboards are popular within the research community as they provide a quick overview of the latest top-performing models and their rankings. Model performance is usually presented by a quantitative metric such as accuracy, F1 score, BLEU, or ROUGE depending on the benchmark/task under assessment, whereas model rank represents the position of the model compared to others on the same benchmark/task.

LLM leaderboards present a common ground for comparing various models against the same benchmarks across different skill sets, such as machine translation, summarization, and question answering. For instance, the Open LLM Leaderboard ~\cite{hgmain} was one of the early leaderboards, with the objective of enabling the evaluation of any model under the exact same setup (same questions, same order, etc.). It covers benchmarks such as ARC~\cite{clark2018think}, MMLU~\cite{hendrycks2020measuring}, HellaSwag~\cite{zellers2019hellaswag}, and TruthfulQA~\cite{lin2021truthfulqa}. Another example is the Holistic Evaluation of Language Models (HELM)~\cite{helm}, which is composed of a set of leaderboards, each focusing on specific capabilities, such as Helm-Safety~\cite{helmsafety}, which focuses on a collection of safety and risk benchmarks (e.g., violence, fraud, discrimination, harassment), VHELM~\cite{vhelm}, which focuses on assessing vision-language models, and MedHELM~\cite{medhelm} for medical tasks.

There are also leaderboards that are based on user votes, such as LMArena, a web-based platform that evaluates LLMs through crowd-sourced, pairwise comparisons from anonymous users. Users enter prompts for two anonymous models and, according to each model’s response, vote for their preferred response. After the vote is cast, the two model identities are revealed. LMArena~\cite{lmsys} was established by Large Model Systems (LMSYS)~\cite{lmsys}, a non-profit organization, and the rankings on the leaderboard are based on an ELO rating system~\cite{elo}. The spread of leaderboards has become more focused on different NLP tasks, such as TTS-Arena~\cite{tts} for text-to-speech models, the Open ASR Leaderboard~\cite{openasr} for speech-to-text models, and the Open VLM Leaderboard~\cite{vlm} for vision-language models.

%\todo{Wouldn't it be better to introduce tasks and the associated leaderboards?} 

%..........................................

\definecolor{permcolor}{HTML}{A8E6CF}
\definecolor{noncommcolor}{HTML}{FFB6C1}
\definecolor{unknowncolor}{HTML}{FFDAB9}
\newcommand{\licensebox}[1]{\textcolor{#1}{\rule{0.7em}{0.7em}}}

\sisetup{group-separator={,}, group-minimum-digits=3}

\section{Existing Medical Leaderboards}
\label{sec:medleaderboards}

% \todo[inline]{WHat about starting with: in recent years, LLMs have increasingly been used for complex analysis tasks in healthcare. In order to compare their performance, domain and tasl-specific leaderboards have evolved ...}

In recent years, many leaderboards have been developed to evaluate LLMs in the medical domain. This section focuses on leaderboards that include multiple benchmarks to comprehensively assess LLM performance. Many of these platforms are openly accessible, enabling both public inspection and reproducibility of results. A notable recent trend is the shift from traditional QA-style evaluations toward more diverse, skill-based assessments. These assessments are designed to test a range of capabilities, such as clinical reasoning, information extraction, and summarization skills, rather than relying only on multiple-choice or short-answer formats. Table~\ref{tab:skillbench} provides an example illustrating how different skill sets can be mapped to corresponding medical benchmarks. This broader evaluation approach has been adopted by several leaderboards, including for instance ClinicBench~\cite{liu2024large} and MedHELM~\cite{bedi2025medhelm} leaderboards, which incorporate varied benchmarks targeting distinct skill categories. The remainder of this section provides an overview of selected medical LLM leaderboards and examines the scope of the benchmarks they cover.

\begin{table*}[!t]
\caption{Examples of Skill Sets and Corresponding Medical Benchmarks}
\label{tab:skillbench}
\centering
\begin{tabular}{p{5cm} p{5cm} p{5cm}}
\toprule
\textbf{Skill Category} & \textbf{Specific Skill} & \textbf{Benchmark Examples} \\
\bottomrule
\textbf{Question \& Answering}  & Multiple-Choice QA & MedQA~\cite{jin2021medqa}, MedMCQA~\cite{pal2022medmcqa}, 
MMLU~\cite{hendrycks2020measuring} \\
\addlinespace
& Open-Ended QA &LiveQA~\cite{abacha2017liveqa}, MedicationQA~\cite{abacha2019bridging} \\
\midrule
\textbf{Reasoning} & Decision Making & PubMedQA~\cite{jin2019pubmedqa} \\
\addlinespace
& Differential Diagnosis &DDXPlus\cite{fansi2022ddxplus} \\
\midrule
\textbf{Vision}
& Question Answering & VQA-RAD~\cite{lau2018dataset} \\
\midrule
\textbf{Summarization}
& Consumer Long Inquiry & MeQSum~\cite{abacha2019summarization} \\
\midrule

\textbf{Reporting} & Note Generation & Aci-bench~\cite{yim2023aci}, CliniKnote~\cite{li2025improving}\\
\addlinespace
\midrule

\textbf{Information Extraction} & Named Entity Recognition (NER) & BC5CDR~\cite{li2016bc5cdr} \\
\addlinespace
& Relation Extraction (RE) & GAD~\cite{bravo2015extraction} \\
\midrule

\textbf{Information Retrieval} & Document Retrieval & TREC Clinical Trials~\cite{treccds2023} \\
\midrule

\textbf{Ethical Decision Making} & Fairness \& Bias & MIMIC-IV~\cite{meng2022interpretability}, EquityMedQA~\cite{pfohl2024toolbox}  \\
\addlinespace
& Safety & Medsafetybench~\cite{han2024medsafetybench} \\
\bottomrule
\end{tabular}
\end{table*}
\subsection{Open Medical-LLM}
The Open Medical-LLM Leaderboard~\cite{hflb} is a board that lists different LLM performances on medical question answering tasks. The main objective is to provide a broad assessment of each model's medical knowledge, medical reasoning capabilities, and question answering abilities~\cite{hflb_doc}. The datasets covered in the leaderboard are MedQA (USMLE), PubMedQA, MedMCQA, and parts of MMLU that relate to medicine and biology. It is open for the research community to submit their models, and the evaluation metric used is accuracy (ACC). Only models that are actively accessible are included in the leaderboard, enabling submission result validation and maintaining the leaderboard's integrity and transparency.
\subsection{MIRAGE}
Medical Information Retrieval-Augmented Generation Evaluation (MIRAGE)~\cite{xiong2024benchmarking,miragelb} is a leaderboard that covers 7,663 questions from five medical QA datasets: MMLU-Med, MedQA-US, MedMCQA, PubMedQA, and BioASQ-Y/N~\cite{mirage}. The core idea behind MIRAGE is to tackle LLMs' hallucination challenges and outdated knowledge by employing the Retrieval-Augmented Generation (RAG) method for the medical domain~\cite{rag}. RAG is a method that improves LLM outputs by combining them with information retrieval systems. Thus, in RAG, the outcome does not rely only on the LLMs' pre-trained knowledge; instead, it accesses and extracts information from external, specific sources, which makes the results more accurate and relevant. The MIRAGE approach utilizes the RAG technique by introducing MedRAG, a toolkit that combines different retrieval methods, corpora, and LLMs~\cite{medrag}. Retrievers used include lexical-based retrievers like BM25~\cite{pyserini,robertson2009probabilistic}, and semantic retrievers such as a general-domain semantic retriever (Contriever)~\cite{contriever,izacard2021unsupervised}, a scientific-domain retriever (SPECTER)~\cite{specter,cohan2020specter}, and a biomedical-domain retriever (MedCPT)~\cite{MedCPT,jin2023medcpt}.

\begin{figure}[!t]
\centering
\hspace*{-0.58cm}\includegraphics[width=1.13\columnwidth]{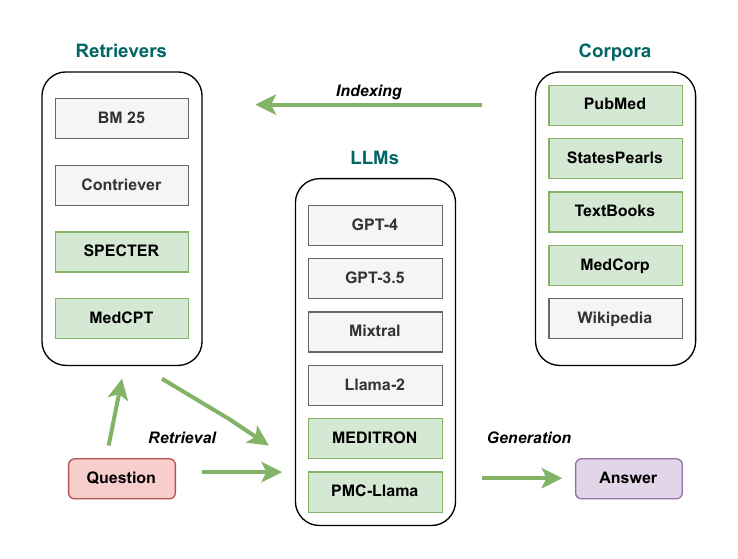}
\caption{MedRAG Toolkit Workflow.}
\label{fig:medragflow}
\end{figure}

For the corpora used in MedRAG, the data is gathered from PubMed (all biomedical abstracts, totaling 23.9 million documents)~\cite{pubmed}, StatPearls~\cite{statpearls} for clinical decision support (covering 9.3k documents and 301.2k snippets), medical textbooks for domain-specific knowledge (covering 18 documents and 125.8k snippets)~\cite{jin2021disease}, and Wikipedia for general knowledge (covering 6.5 million documents and 29.9 million snippets). Lastly, the MedCorp corpus includes 30.4 million documents and 54.2 million snippets. All snippets from these corpora enhance cross-source retrieval for different inquiries. Figure~\ref{fig:medragflow} demonstrates the workflow of the MedRAG toolkit.

\subsection{MedHELM}
\begin{figure*}[!t]
\centering
\includegraphics[width=0.85\textwidth]{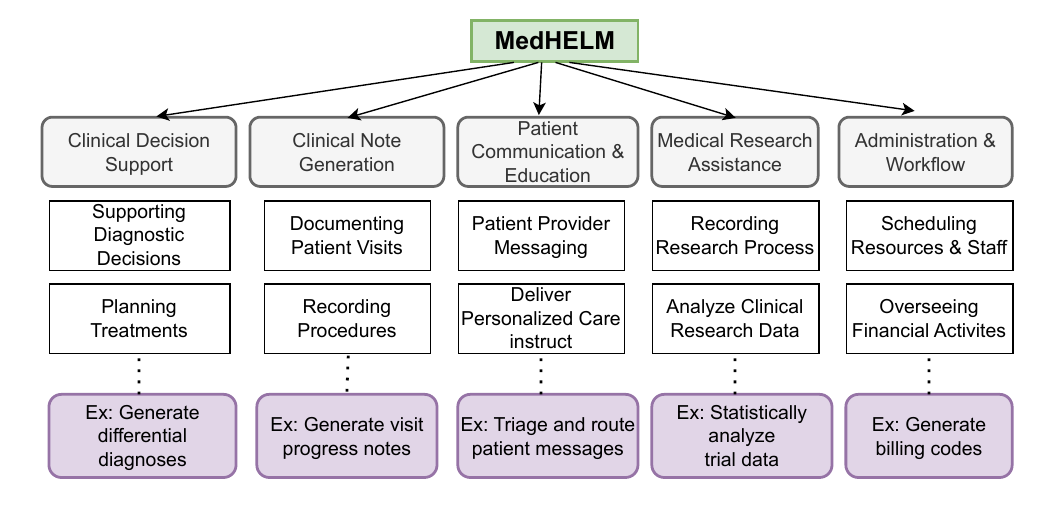}
\caption{Overview of MeHELM Categories, Subcategories and tasks~\cite{bedi2025medhelm}.}
\label{fig:medhelm}
\end{figure*}

MedHELM ~\cite{bedi2025medhelm} is one of the HELM leaderboards, which focuses exclusively on evaluating LLM medical tasks. It consists of 5 categories, 22 subcategories, and 121 clinical tasks covered by 35 benchmarks. The five categories are: Clinical Decision Support; Patient Communication; Clinical Note Generation; Medical Research Assistance; and Administration and Workflow. An example of the taxonomy covered is shown in Figure~\ref{fig:medhelm}. The objective of the MedHELM leaderboard is to conduct evaluations of practical applications of language models in healthcare. Unlike the Open LLM Leaderboard, HELM is not open for public submissions. It is maintained by the Stanford Center for Research on Foundation Models (CRFM), which evaluates predefined state-of-the-art (SOTA) models from major research labs and companies. The 35 benchmarks used are distributed as follows: 14 private, 7 gated-access (i.e., requiring approval), and 14 public. MedHELM collaborated with various medical and technology institutions to build the leaderboard, which brought diverse coverage in the benchmarks used. It also covers a set of LLMs from some of the pioneers in the field. While MedHELM provides open documentation to facilitate leaderboard reproduction, some benchmarks remain private because they are proprietary or restricted datasets available only to the respective organizations~\cite{bedi2025medhelm}. Closed leaderboards may be discouraging for the broader research community, as smaller labs or open-source projects might struggle to gain fair recognition there.

Several other medical leaderboards target different aspects of the domain. ClinicBench~\cite{liu2024large} supports a diverse set of benchmarks across various medical tasks~\cite{clinicbench1,clinicbench2}. MMedBench~\cite{qiu2024towards} focuses on medical multiple-choice questions spanning six languages~\cite{multilingmedqa}. HealthBench focuses on realistic health conversations, each graded using custom, physician-created criteria~\cite{arora2025healthbench}.

\section{Medical Benchmarks Used in Leaderboards}
\label{sec:benchmarks}
In this section, we explore a selected set of medical benchmarks that are commonly used on leaderboards, as summarized in Table~\ref{tab:leadbench}, also used in evaluating many models like MedPrompt~\cite{nori2023can}, GPT4~\cite{nori2023capabilities}, Med\-PaLM~\cite{singhal2025toward}, Claude 3~\cite{TheC3} and MEDITRON~\cite{chen2023meditron}, summarized in Table~\ref{tab:modelbench}. By analyzing these benchmarks, we aim to uncover how leaderboard rankings translate into measurable capabilities and identify the specific skills they evaluate, thereby helping to build additional benchmarks that address any gaps in assessing medical models.

\begin{table}[t]
    \caption{Medical Benchmarks Used in Various Leaderboards}
    \label{tab:leadbench}
    \centering
    \renewcommand{\arraystretch}{1.2}
    \setlength{\tabcolsep}{4pt} % reduce column spacing
    \resizebox{\columnwidth}{!}{ % shrink to column width 
    \begin{tabular}{lcccc}
        \toprule
        \textbf{Benchmark} & \textbf{Open Medical-LLM} & \textbf{MIRAGE} & \textbf{MedHELM} & \textbf{ClinicBench} \\
        \midrule
        MedQA       & $\checkmark$ & $\checkmark$ & $\times$ & $\checkmark$ \\
        MMLU        & $\checkmark$ & $\checkmark$ & $\times$ & $\checkmark$ \\
        MedMCQA     & $\checkmark$ & $\checkmark$ & $\times$ & $\checkmark$ \\
        PubMedQA    & $\checkmark$ & $\checkmark$ & $\checkmark$ & $\checkmark$ \\
        BIOASQ      & $\times$     & $\checkmark$ & $\times$ & $\times$ \\
        \bottomrule
    \end{tabular}}
\end{table}

\begin{table}[t]
    \caption{Medical Benchmarks Used in Evaluation for Various Models}
    \label{tab:modelbench}
    \centering
    \renewcommand{\arraystretch}{1.2}
    \setlength{\tabcolsep}{4pt} % reduce column spacing
    \resizebox{\columnwidth}{!}{ % shrink to column width
    \begin{tabular}{l p{2cm} p{1.7cm} p{2.2cm} p{1.5cm} p{1.7cm}}
        \toprule
        \textbf{Benchmark} & \textbf{MedPrompt} & \textbf{GPT4} & \textbf{Med-PaLM2} & \textbf{Claude3} & \textbf{MEDITRON}\\
        \midrule
        MedQA       & $\checkmark$ & $\checkmark$ & $\checkmark$ &$\checkmark$ & $\checkmark$ \\
        MMLU        & $\checkmark$ & $\checkmark$&$\checkmark$ &  $\times$ & $\checkmark$ \\
        MedMCQA     & $\checkmark$ & $\checkmark$&$\checkmark$ &  $\checkmark$ & $\checkmark$ \\
        PubMedQA    & $\checkmark$ & $\checkmark$&$\checkmark$ &  $\checkmark$ & $\checkmark$ \\
        BIOASQ      & $\times$  & $\times$ &$\times$     & $\checkmark$ & $\times$ \\
        \bottomrule
    \end{tabular}}
\end{table}

\subsection{MedQA}
The MedQA dataset benchmark~\cite{jin2021medqa} is considered one of the first open-domain QA (OpenQA) datasets aimed at finding answers to medical questions based on large-scale information sources such as search engines or Wikipedia. Examples of OpenQA datasets include ARC~\cite{clark2018think} and OpenBookQA~\cite{mihaylov2018can}. MedQA was developed from professional national board exams in the USA, 
% \todo{ARe these names politically correct?} 
China, and Taiwan, known in the literature as USMLE, MCMLE, and TWMLE, respectively. MedQA covers three main languages: English with 12,723 questions, Simplified Chinese with 34,251 questions, and Traditional Chinese with 14,123 questions. The main objective of these exams is to evaluate doctors’ ability to apply their acquired knowledge, understand concepts, and demonstrate decision-making skills. The exam questions typically begin by describing a medical use case involving a patient’s condition, followed by multiple-choice answers targeting diagnosis or treatment options.
\subsection{MMLU}
MMLU (Massive Multitask Language Understanding)~\cite{hendrycks2020measuring} is a dataset designed to assess the knowledge acquired by LLMs during pretraining. It contains a medical part covering topics like -clinical knowledge, professional medicine, anatomy and more. It evaluates models in both zero-shot and few-shot settings. The benchmark aims to align more closely with human evaluation by including a wide range of difficulty levels, from elementary to advanced professional, and by assessing both world knowledge and problem-solving abilities. MMLU includes 15,908 multiple-choice questions~\cite{mmlugit}, covering 57 subjects across STEM, the humanities, the social sciences, and other domains.
\subsection{MedMCQA}
The MedMCQA dataset is a large-scale multiple-choice QA benchmark~\cite{pal2022medmcqa}. It consists of more than 194K questions extracted from two medical entrance exams: the All India Institute of Medical Sciences (AIIMS PG) and the National Eligibility cum Entrance Test (NEET PG). The primary objective of these exams is to evaluate doctors’ ability to apply their acquired knowledge and use reasoning to diagnose presented cases and make appropriate treatment decisions. MedMCQA includes several medical topics such as Anesthesia, Anatomy, Biochemistry, Dentistry, Medicine, and others. The exams cover approximately 2.4K healthcare topics and 21 medical subjects. In MedMCQA, each question presents a medical query associated with four possible answer choices, some of which include an explanation section. Such explanations can be used as contextual information, as referenced by the leaderboard~\cite{medmcqagit}. The testing subset of MedMCQA comprises 4,183 records.
\subsection{PubMedQA}
PubMedQA~\cite{jin2019pubmedqa} is a healthcare and biomedical question-answering (QA) dataset compiled from PubMed abstracts and articles. The objective of PubMedQA is to assess system capability in reasoning by answering medical research questions with one of three possible responses: yes, no, or maybe (e.g., \textit{Do mitochondria play a role in remodeling lace plant leaves during programmed cell death?}). The answer depends on the corresponding article abstract. The benchmark leaderboard~\cite{pubmedqalb} is based on a 500-question expert-annotated test set. PubMedQA also released a 1,000-question expert-annotated dataset named \textit{PQA-L(Labeled)}. Additionally, it provides \textit{PQA-U(Unlabeled)}, a collection of 61.2K PubMed articles without labeled answers, and \textit{PQA-A(Artificial)}, a collection of 211.3K PubMed articles that include artificially generated labeled answers.
\subsection{BIOASQ}
BioASQ is an annual large-scale competition in biomedical semantic indexing and question answering (QA). It has been running since 2013, with a focus on assessing the capability of systems to semantically index large-scale collections of medical scientific articles and make them available for natural language QA systems, enabling them to return relevant answers to users. Two tasks were initially in scope. The first was the automatic classification of biomedical documents using concepts from knowledge bases (e.g., Medical Subject Headings). The second task involved answering biomedical questions by extracting text fragments from scientific articles~\cite{tsatsaronis2015overview}. The competition has expanded over the past decade and currently covers diverse tasks such as biomedical text classification, information retrieval, machine learning, QA from texts and structured data, and multi-document summarization~\cite{nentidis2023overview,nentidis-etal-2018-results,nentidis2021overview}. There is no single standard leaderboard as in the Open Medical Leaderboard and MIRAGE; instead, there are detailed results pages listing the competition outcomes for each year, along with additional subcategories based on that year’s tasks~\cite{bioasqresults}.

% ...............................................

\section{Beyond Leaderboard and Benchmarks}
\label{sec:lbdiscussion}

In the previous sections, we have introduced and exemplified leaderboards and benchmarks for the assessment of LLMs in the healthcare domain. However, the main findings are independent of this domain and can be transferred and generalized to other domains:

\emph{Benchmark Contamination:} Dataset or benchmark contamination occurs when LLMs unintentionally include evaluation benchmark information in their training or fine-tuning data, leading to inaccurate, false, or unreliable results during model evaluation. A possible solution is to move toward private or gated benchmarks, or to use dynamic benchmarks with continuously refreshed content~\cite{xu2024benchmark,hasan2025pitfalls}.

\emph{Benchmark Saturation:} Some benchmarks are no longer challenging and have become too easy, reaching baseline human performance. This has occurred in the Open LLM Leaderboard, leading to a refactoring of the leaderboard to replace those benchmarks~\cite{opnllmblog}.

\emph{Benchmark Selection Bias:} Deciding which benchmarks to include can be subjective, as leaderboards may misrepresent the practicality and usefulness of certain benchmarks for real-world applications, especially in domains such as medicine. The continuous participation of domain-specific subject matter experts is therefore essential~\cite{dehghani2021benchmark}.

\emph{Computation and Maintenance Cost:} Running an evaluation for a model requires significant computational resources, which explains why current leaderboards are typically supported by large organizations. Maintaining and keeping leaderboards relevant requires continuous updates to benchmarks, evaluation methods, and infrastructure. Actively updated leaderboards consistently gain the community’s trust.

\emph{Evaluation Metrics:} Many leaderboards rely on one or two scores, such as accuracy, average performance, or efficiency (e.g., inference time in seconds), which may oversimplify model capabilities. A model might perform very well on certain tasks but poorly on others, yet aggregation hides these weaknesses. Furthermore, automated statistical metrics such as BLEU, ROUGE, or exact match do not effectively measure reasoning quality, truthfulness, or usefulness of outputs. Some leaderboards have addressed this by using LLMs as judges; while widely adopted, this approach is still under investigation~\cite{chen2024evaluating,wang2025survey,alzahrani2024benchmarks}.

\emph{Overfitting in Arena-type Leaderboards:}
For leaderboards based on Elo ratings and user votes, such as Chatbot Arena, certain regulations are needed to prevent overfitting, as detailed in \textit{The Leaderboard Illusion}~\cite{singh2025leaderboard}. Some of the recommendations include: prohibiting score withdrawal after submission, setting transparent limits on the number of private model variants per provider, establishing clear and auditable model deprecation criteria, and enhance sampling fairness.

\begin{table}[t]
\centering
\setlength{\tabcolsep}{3pt}
\caption{Top downloaded Hugging Face LLMs (snapshot: 30 Sep 2025).}
\label{tab:hf_top_models}
\begin{tabular}{>{\centering\arraybackslash}p{0.6cm}   
    >{\raggedright\arraybackslash}p{3.1cm} 
    >{\raggedleft\arraybackslash}p{1.5cm}  
    >{\raggedleft\arraybackslash}p{1.3cm}  
    >{\raggedleft\arraybackslash}p{1.1cm}  
    }
\toprule
\textbf{L} & \textbf{Model} & \textbf{Downloads} & \textbf{Likes} & \textbf{Size} \\
\midrule
\licensebox{permcolor} & openai-community/gpt2 & \textbf{\num{11567187}} & \num{2958} & 137M \\
\licensebox{noncommcolor} & facebook/opt-125m & \num{8105841} & \num{218} & 125M \\
\licensebox{noncommcolor} & Qwen/Qwen2.5-3B-Instruct & \num{8073896} & \num{311} & 3.0B \\
\licensebox{permcolor} & meta-llama/Llama-3.1-8B-Instruct & \num{7192139} & \num{4689} & 8.0B \\
\licensebox{permcolor} & Qwen/Qwen2.5-7B-Instruct & \num{7184004} & \num{806} & 7.0B \\
\licensebox{permcolor} & meta-llama/Llama-3.2-1B-Instruct & \num{6849770} & \num{1087} & 1.0B \\
\licensebox{permcolor} & Qwen/Qwen3-0.6B & \num{6748616} & \num{652} & 600M \\
\licensebox{permcolor} & openai/gpt-oss-20b & \num{6637334} & \num{3632} & 20.0B \\
\licensebox{permcolor} & Qwen/Qwen3-32B & \num{5660241} & \num{541} & 32.0B \\
\licensebox{permcolor} & google/gemma-3-1b-it & \num{5636363} & \num{633} & 1.0B \\

\bottomrule
\end{tabular}

\vspace{0.3em}
\raggedright
Legend:
\licensebox{permcolor} Permissive \quad
\licensebox{noncommcolor} Non-comm. 
\end{table}

\section{Community Endorsement}
\label{sec:commu}
Besides navigating leaderboards, there are other ways to gain insights into model performance. This can be done by obtaining insights from platforms like Hugging Face~\cite{hgwiki}, the most popular platform that provides access to thousands of pre-trained models, datasets, and libraries. It enables developers and researchers to build, share, and deploy AI models easily. Hugging Face provides filtering options for users to navigate models and datasets, for example, sorting the models by top downloads, likes, or trending status. Downloads, for instance, can reflect how often a model has been pulled, offering insights into its popularity and adoption. Thus, the higher the number of downloads, the stronger the indication that such models are widely used and trusted by the community. It can suggest that the model is well-documented, easy to integrate, or reliable across tasks. However, downloads can be inflated by automated pipelines or repeated pulls, so they may not always measure user satisfaction or quality.
Sorting models by likes in Hugging Face can also shed light on community approval and endorsement of a model. The more likes a model has, the more indicative it is that users appreciate its quality, reliability, usefulness, ease of use, innovative architecture, or good documentation. Unlike downloads, likes do not necessarily indicate usage volume. A niche model could have fewer downloads but a high like-to-download ratio if the people who tried it liked it a lot. Nevertheless, likes can also be subjective, someone could like a model for its documentation, speed, or even novelty, not purely for its accuracy or performance.
Therefore, to build a more confident understanding of current models, it is better to consider likes along with downloads, trends, and evaluation metrics on benchmarks before choosing a model. We have extracted a snapshot (covering 30 days) from Hugging Face to view the top downloaded and liked models for the text-generation task, as shown in Table~\ref{tab:hf_top_models} and Table~\ref{tab:hf_liked_models}, respectively, and we noticed the following:
\begin{itemize}
\item The models that are highly downloaded and highly liked are mostly in the permissive license spectrum. We define permissive in this paper as a license that allows usage in both commercial and research settings (though some licenses have detailed restrictions), such as Apache 2.0, MIT, Gemma, Llama 2, and Llama 3.
\item The models that are highly downloaded usually favor small models in terms of parameter size. For Table~\ref{tab:hf_top_models}, the range of models is mostly under 3B parameters, which is expected, as such models are more affordable to try and explore.
\item When comparing the content of Table~\ref{tab:hf_top_models} versus Table~\ref{tab:hf_liked_models}, the models that are highly downloaded are not necessarily the most liked models, which encourages maintaining more options in the model selection decision.
\item The models that are highly downloaded and/or liked are dominated by the leading organizations in the field, which is expected given the cost of pretraining these models from scratch.
\end{itemize}
These insights can motivate the selection process, even if only for one task (text generation), but they do not necessarily imply that the most downloaded or liked model is the best. The process of selecting the right model for a particular use case is sophisticated and customizable, depending on many other factors, as will be detailed in Section~\ref{sec:msm}.

\begin{table}[t]
\centering
\setlength{\tabcolsep}{3pt}
\caption{Top Hugging Face LLMs by user likes (snapshot: 30 Sep 2025).}
\label{tab:hf_liked_models}
\begin{tabular}{>{\centering\arraybackslash}p{0.6cm}   
    >{\raggedright\arraybackslash}p{3.1cm} 
    >{\raggedleft\arraybackslash}p{1.5cm}  
    >{\raggedleft\arraybackslash}p{1.3cm}  
    >{\raggedleft\arraybackslash}p{1.1cm}  
    }
\toprule
\textbf{L} & \textbf{Model} & \textbf{Downloads} & \textbf{Likes} & \textbf{Size} \\
\midrule
\licensebox{permcolor} & meta-llama/Meta-Llama-3-8B & \num{2120284} & \textbf{\num{6332}} & 8.0B \\
\licensebox{permcolor} & meta-llama/Llama-3.1-8B-Instruct & \num{7192139} & \num{4689} & 8.0B \\
\licensebox{permcolor} & openai/gpt-oss-120b & \num{3572640} & \num{3912} & 120.0B \\
\licensebox{permcolor} & openai/gpt-oss-20b & \num{6637334} & \num{3632} & 20.0B \\
\licensebox{permcolor} & openai-community/gpt2 & \num{11567187} & \num{2958} & 137M \\
\licensebox{permcolor} & meta-llama/Llama-2-7b-hf & \num{2133119} & \num{2158} & 7.0B \\
\licensebox{permcolor} & meta-llama/Llama-3.2-1B & \num{2293832} & \num{2093} & 1.0B \\
\licensebox{permcolor} & meta-llama/Llama-3.2-3B-Instruct & \num{1832773} & \num{1739} & 3.0B \\
\licensebox{permcolor} & deepseek-ai/DeepSeek-R1-Distill-Qwen-32B & \num{3207520} & \num{1446} & 32.0B \\
\licensebox{permcolor} & TinyLlama/TinyLlama-1.1B-Chat-v1.0 & \num{2404466} & \num{1416} & 1.1B \\

\bottomrule
\end{tabular}

\vspace{0.3em}
\raggedright
Legend:
\licensebox{permcolor} Permissive \quad
\licensebox{noncommcolor} Non-comm. 
\end{table}

\section{Model Selection Methodology (MSM)}
\label{sec:msm}

\begin{figure*}[!t]
\centering
\includegraphics[width=1\textwidth, height=0.15\textheight]{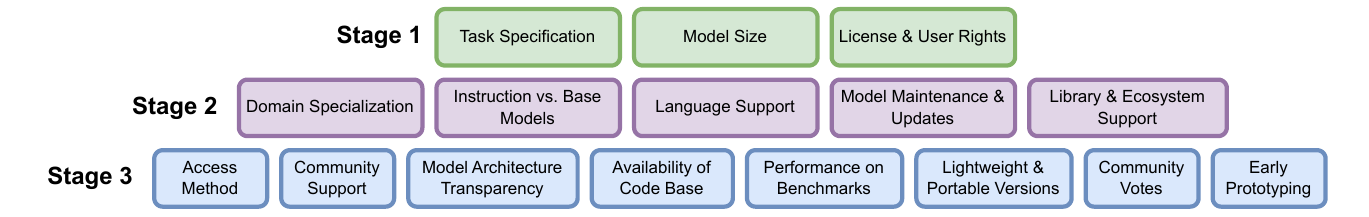}
\vspace{-0.5em}
\caption{Model Selection Methodology}
\label{fig:msm}
\end{figure*}

It is clear that LLMs have become integral to everyday life in recent years. Market research has anticipated rapid growth in the LLM sector. For instance, Grand View Research estimated the global market for LLMs at USD 5.61 billion in 2024, with a projected compound annual growth rate (CAGR) of 36.9\% from 2025 to 2030~\cite{llmmarket,limonad2025selecting}. 
This underscores the increasing importance of selecting the appropriate model, and highlights the need for a more structured selection process for practitioners. While leaderboards and benchmarks provide useful guidance in comparing models for research or business applications, other factors are equally important in narrowing down choices. To address this, we present the MSM (Model Selection Methodology), an intuitive methodology for navigating different SOTA models and guiding the model decision-making process.
MSM is organized into three stages, reflecting different types of considerations, as illustrated in Figure~\ref{fig:msm}. The earlier stages focus on critical requirements, while the later stages address important factors and contextual validation. Although MSM is represented as a three-stage pyramid, these stages should not be interpreted as strict levels of importance. MSM is inherently contextual, and some criteria may shift in criticality or relevance depending on the specific use case.

\subsection{Stage 1: (Critical)}
This stage defines the prerequisite conditions. If these conditions are not met, it will likely be necessary to build and pretrain a model from scratch or adjust the project requirements significantly.
\subsubsection{Task Specification}
The first step is to specify the objective task (e.g., text generation, text-to-speech, image-to-text, etc.). Hugging Face lists over 40 tasks, which can be filtered to identify relevant models as seen in Table~\ref{tab:hftasks}. This step narrows the scope to what is available and avoids considering irrelevant use cases.

\begin{table*}[!t]
\caption{Examples of Tasks Listed in Huggingface Platform}
\label{tab:hftasks}
\centering
\begin{tabular}{p{4cm} p{12cm}}
\toprule
\textbf{Task Category} & \textbf{Tasks} \\
\midrule
\textbf{Multimodal}  & 
Audio-Text-to-Text \textbullet\ Image-Text-to-Text \textbullet\ Visual Question Answering \textbullet\ Document Question Answering \textbullet\ Video-Text-to-Text \textbullet\ Visual Document Retrieval \textbullet\ Any-to-Any \\
\midrule
\textbf{Computer Vision} & 
Depth Estimation \textbullet\ Image Classification \textbullet\ Object Detection \textbullet\ Image Segmentation \textbullet\ Text-to-Image \textbullet\ Image-to-Text \textbullet\ Image-to-Image \textbullet\ Image-to-Video \textbullet\ Unconditional Image Generation \textbullet\ Video Classification \textbullet\ Text-to-Video \textbullet\ Zero-Shot Image Classification \textbullet\ Mask Generation \textbullet\ Zero-Shot Object Detection \textbullet\ Text-to-3D \textbullet\ Image-to-3D \textbullet\ Image Feature Extraction \textbullet\ Keypoint Detection \textbullet\ Video-to-Video \\
\midrule
\textbf{Natural Language Processing} &
Text Classification \textbullet\ Token Classification \textbullet\ Table Question Answering \textbullet\ Question Answering \textbullet\ Zero-Shot Classification \textbullet\ Translation \textbullet\ Summarization \textbullet\ Feature Extraction \textbullet\ Text Generation \textbullet\ Fill-Mask \textbullet\ Sentence Similarity \textbullet\ Text Ranking \\
\midrule
\textbf{Audio} &
Text-to-Speech \textbullet\ Text-to-Audio \textbullet\ Automatic Speech Recognition \textbullet\ Audio-to-Audio \textbullet\ Audio Classification \textbullet\ Voice Activity Detection \\
\midrule
\textbf{Tabular} & 
Tabular Classification \textbullet\ Tabular Regression \textbullet\ Time Series Forecasting \\
\midrule
\textbf{Reinforcement Learning} & 
Reinforcement Learning \textbullet\ Robotics \\
\bottomrule
\end{tabular}
\end{table*}

\subsubsection{Model Size and Computational Cost}
Model size directly impacts training and inference costs, hardware requirements, and latency. Many institutions are promoting compact yet high-performing models. Model size is so central that it is often embedded in model names. Additionally, smaller models are generally more energy efficient and are downloaded more frequently, reflecting practical usability~\cite{chen2024role,abdin2024phi, phi}.
\subsubsection{License and User Rights}
Licensing defines whether a model can be used commercially or only for research. It also constrains whether modifications, redistribution, or integration with proprietary systems are allowed. Well-known licenses include MIT, Creative Commons, and Apache~2, while specialized licenses such as LLaMA~2 and Gemma impose stricter conditions. This factor requires careful alignment with the intended use case. Performing an assessment of bias, fairness, interpretability, and data privacy is also important. Even technically strong models may be unsuitable if they introduce unacceptable risks or fail to meet compliance requirements~\cite{hfmodel,variouslic,opensourcelic}.
\subsection{Stage 2: (Important)}
This stage covers secondary yet important conditions. If these conditions are unmet, they may not prevent model use entirely, but they will require additional adjustments, such as fine-tuning, adaptation, or further development to close the gaps.

\subsubsection{Domain Specialization}
Certain applications demand domain adaptation (e.g., biomedical, finance, telecom)~\cite{zhao2023beyond}. Several ready-made domain-adapted models are available. For example, in the medical domain~\cite{wu2024pmc,labrak2024biomistral,lee2020biobert}, selecting a medical foundation model can considerably accelerate the development of medical use cases compared to training from a general-purpose model~\cite{tejani2022performance}.

\subsubsection{Instruction vs. Base Models}
Models are often released in two variants: base and instruction-tuned. Base models serve as general-purpose backbones but usually require fine-tuning or prompting strategies to perform well on downstream tasks. Instruction-tuned models, on the other hand, are aligned to follow natural language instructions, enabling more effective out of the box use in many applications. Selecting the appropriate variant can substantially reduce adaptation effort and training cost, depending on the use case.
\subsubsection{Language Support}
LLM models usually come in a multilingual setting covering several languages. Though English and many other languages are widely adopted in current models, the lack of certain languages relevant to the target use case or project may require costly pretraining. For instance, it may take a reasonable amount of time to pretrain a small model in a new language with appropriate hardware, whereas larger models require a far greater investment. Additionally, to adapt the model to the missing language without compromising support for currently supported languages, data preparation must meet adequate quality and quantity standards~\cite{yong2025state,nguyen2023democratizing,kiulian2024english}.
\subsubsection{Library and Ecosystem Support}
Modern model development relies on ecosystems (e.g., Hugging Face, DeepSpeed, PyTorch Lightning). Models with broader library support enable smoother training, optimization, distributed computing, and deployment pipelines. Ecosystem integration also includes support for ONNX, TensorRT, or cloud deployment, which can significantly reduce engineering overhead.

\subsubsection{Model Maintenance and Updates}
The long-term reliability of a model depends on whether it is actively maintained, patched, and updated. A language model is not just a large blob of weights; it is part of a broader software stack including frameworks, loaders, and deployment infrastructure that must remain secure. Libraries or frameworks (e.g., PyTorch, Transformers) that are not regularly updated may contain vulnerabilities that attackers can exploit~\cite{harzevili2023characterizing}. Models can also pose risks related to data privacy violations, as outdated models may retain or unintentionally reveal sensitive training data if they were not fine-tuned, filtered, or audited under modern privacy and compliance standards (e.g., GDPR, HIPAA)~\cite{gdpr,hipaa}.
The research community is actively studying emerging privacy and security flaws that may affect language models, such as model poisoning~\cite{souly2025poisoning}, prompt injection~\cite{liu2024automatic,yan2024backdooring}, and other evolving safety concerns~\cite{shi2024large}. These developments reinforce the importance of selecting models that are actively maintained to ensure a more stable, secure, and compliant deployment ecosystem.

\subsection{Stage 3: (Contextual Validation)}
This stage involves contextual validation activities that do not necessarily disqualify a model but provide valuable insights for more informed decision-making. It is positioned last not because it is optional, but because it can only be meaningfully performed after foundational constraints and selection factors have been addressed. For example, this stage incorporates \textit{Early Prototyping}, an important step that evaluates shortlisted models in the actual application setting to provide concrete, evidence-driven feedback.

\subsubsection{Access Method}
Models may be available via direct download (requiring dedicated hardware) or API access (subject to usage limits, costs, and latency). The choice of access method impacts scalability and planning for throughput and token consumption.

\subsubsection{Community Support}
An active online community can accelerate adoption through shared experiences, tutorials, and open discussions. This improves troubleshooting and accelerates innovation.

\subsubsection{Model Architecture Transparency}
Availability of the model’s technical reports, research papers, documentation, training datasets, and evaluation methods provides insight into its performance and potential limitations. Transparency also aids in fostering innovation and adaptation

\subsubsection{Availability of Code Base}
Access to the source code for model architectures or training pipelines enhances reproducibility and innovation. In addition, examining repository metadata for instance, on GitHub this could include the number of \texttt{stars}, \texttt{watch},  \texttt{forks}, \texttt{contributors}, recent code commits, or active discussions, all can serve as informal trust indicators of how active and well-maintained the code base is.

\subsubsection{Performance on Benchmarks}
Benchmark results and leaderboard rankings remain key indicators of model strength. They help researchers identify models excelling in specific tasks and provide data formats for reproducibility. However, reliance only on benchmarks may overlook robustness and real-world generalization.

\subsubsection{Community votes}
Community usage patterns, such as model downloads, likes and trends on platforms like Hugging Face can serve as vote for trust for some models. They highlight widely adopted and practically useful models.
\subsubsection{Lightweight and Portable Versions}
Availability of optimized forms like distilled or quantized model or availability of model exported in ONNX format facilitates deployment on resource-constrained hardware. This extends accessibility and exploring without requiring full-scale infrastructure.

\subsubsection{Early prototyping}
Examining the shortlisted models is important, as by this stage most of the other points to address have already been covered. By examining the model’s behavior according to your use case and using your dataset, you can build a more intuitive understanding of how well the model can be adapted and how far it currently is from the target behavior. This stage can also be beneficial for assessing the model’s readiness within the current use case or project infrastructure and for addressing potential issues early on.

% \todo{Add discussion Section here} 
% ..........................................
\section{Discussion}
\label{sec:discussion}
Selecting the right model for a specific task is a multidimensional decision-making process. As discussed earlier, although public leaderboards and benchmarks offer accessible, quantitative indicators of model performance, they do not capture the full complexity involved in real-world adoption. Benchmark results are often narrow in scope, limited to particular datasets or tasks, and may not reflect constraints related to use-case requirements, cost, latency, security, privacy, maintainability, or deployment infrastructure. As a result, relying solely on leaderboard or benchmark performance is still insightful, yet incomplete for informed model selection, especially in high-stakes or resource-constrained environments.
Historically, the data science and machine learning community has consistently addressed similar forms of complexity by developing standards and structured methodologies that transform unstructured, ad hoc processes into well-defined, repeatable practices. Early efforts such as CRISP-DM (Cross-Industry Standard Process for Data Mining)~\cite{shearer2000crisp} provided the first widely adopted framework for organizing the data-mining lifecycle by proposing six major stages: Business Understanding, Data Understanding, Data Preparation, Modeling, Evaluation, and Deployment. This structure enabled practitioners to translate experience into a systematic process that became the de facto standard for data mining. SEMMA, introduced by SAS, followed a similar spirit by formalizing its iterative steps: Sample, Explore, Modify, Model, and Assess into a repeatable workflow~\cite{azevedo2008kdd}. These initiatives demonstrated the value of codifying practitioner intuition into structured methodologies that could guide teams, increase productivity, and reduce ambiguity.
As the field evolved, principles like FAIR (Findable, Accessible, Interoperable, Reusable)~\cite{wilkinson2016fair} reframed data management around transparency and long-term usability. Similarly, the DAMA-DMBOK framework established comprehensive best practices for data governance, reflecting the growing need to standardize organization-wide data procedures~\cite{international2017dama,dama}. Just as data mining and data management matured through frameworks such as CRISP-DM and DAMA-DMBOK, the discipline of project management is another example that underwent a similar transformation. Historically, project execution differed dramatically across industries and organizations, often relying on individual experience rather than shared structure. The emergence of project-management standards such as PMBOK~\cite{project2021guide}, PRINCE2~\cite{prince2}, and Agile methodologies~\cite{abrahamsson2017agile} illustrated the necessity of codifying best practices into transparent, repeatable processes. 
More recently, the development of MLOps frameworks~\cite{kreuzberger2023machine} and Model Cards~\cite{mitchell2019model} has further pushed the community toward explicit documentation, governance, and operational consistency for machine learning systems. These contributions illustrate a continuous ideology: when the complexity of data and models increases, the community responds by creating frameworks that formalize shared understanding into actionable standards. The success of all these frameworks were not only from process standardization but also from their ability to pin down a shared terminology that practitioners could adopt. Establishing such common language is equally essential in the rapidly evolving LLM ecosystem.
In the context of large language models, the landscape remains highly agile and fragmented. New models appear frequently, architectural innovations evolve rapidly, and performance depends on operational factors that extend far beyond accuracy metrics. Despite this, there is no unified process for navigating the trade-offs involved in LLM model selection. This gap has left practitioners without a structured methodology for evaluating models in a holistic and repeatable manner.
The Model Selection Methodology (MSM) proposed in this work aims to serve as an early step toward standardizing the model selection process for language models. MSM codifies common-sense reasoning (currently practiced informally by engineers and researchers) into a clear, structured workflow that prioritizes requirements, constraints, and context-specific trade-offs. While MSM is not presented as a definitive or exhaustive standard, it provides a foundational blueprint that can be extended, refined, and validated by the broader community. As prior frameworks in data science have shown, early conceptual methodologies often act as seeds that mature into widely accepted standards through iterative research and practical adoption.

\section{Conclusion}
\label{sec:conc}
In this work, we focused on exploring the landscape of leaderboards in general and, in particular, examined the healthcare and medicine domain as a case study. By delving deeper into this domain, we identified several challenges associated with relying solely on leaderboards when selecting a model for a specific use case. We also discussed the general limitations of leaderboards and proposed possible solutions for some of these challenges.
We acknowledge that leaderboards still represent a valuable approach, providing a quick quantitative dimension for a large number of models, as they make it easier to compare various models using the same framework. Nevertheless, leaderboards should not be the only method for model selection. Therefore, we also focused on examining other factors that reflect the community’s preferences and trust toward different models.
Based on these insights, we derived the Model Selection Methodology (MSM), a method that provides a step-by-step approach for navigating, sorting, and shortlisting models for a particular use case or domain. We hope that this methodology can serve as a step toward establishing a more systematic and accelerated model selection process.

\section*{Acknowledgments}
This work was partly supported by the Freiwillige Akademische Gesellschaft (FAG)\footnote{https://fag-basel.ch/} Basel.
% We would like to extend our sincere gratitude for the financial support provided by 
%\censor{the Freiwillige Akademische} %\censor{Gesellschaft (FAG)\footnote{https://fag-%basel.ch/}}.

\bibliographystyle{IEEEtran}
\bibliography{medbenchmark}

\end{document}